\documentclass{article} 
\usepackage{iclr2022_conference,times}

\usepackage{amsmath,amsfonts,bm}

\def\eqref#1{equation~\ref{#1}}

\def\1{\bm{1}}

\DeclareMathAlphabet{\mathsfit}{\encodingdefault}{\sfdefault}{m}{sl}
\SetMathAlphabet{\mathsfit}{bold}{\encodingdefault}{\sfdefault}{bx}{n}

\usepackage{hyperref}
\usepackage[all]{hypcap}
\usepackage{url}
\usepackage[binary-units=true]{siunitx}
\usepackage{graphicx}
\usepackage{caption}
\graphicspath{./figures/}

\newcommand{\fref}[1]{Figure~\ref{#1}}   
\newcommand{\cref}[1]{Chapter~\ref{#1}}  
\newcommand{\sref}[1]{Section~\ref{#1}}  

\title{Trainable Wavelet Neural Network\\for Non-Stationary Signals}

\author{Jason Stock \& Charles Anderson \\
Department of Computer Science\\
Colorado State University\\
Fort Collins, CO, USA \\
\texttt{\{stock,anderson\}@colostate.edu}
}

\iclrfinalcopy 
\begin{document}

\maketitle

\begin{abstract}
This work introduces a wavelet neural network to learn a filter-bank specialized to fit non-stationary signals and improve interpretability and performance for digital signal processing. The network uses a wavelet transform as the first layer of a neural network where the convolution is a parameterized function of the complex Morlet wavelet. Experimental results, on both simplified data and atmospheric gravity waves, show the network is quick to converge, generalizes well on noisy data, and outperforms standard network architectures. 
\end{abstract}

\section{Introduction}

Classical filters in signal processing are excellent tools to transform time-series data in the spatial and frequency domains, based on the type of transformation and filter. For 1D signals, wavelets are often used to cut data into frequency components to localize spectral information. When combined with neural networks, wavelets are often used during preprocessing with a pre-defined filter-bank to use magnitude spectogram features as input. \cite{Lee2019-gs} explore this approach for electroencephalograph classification, \cite{Zhao2019-te} for decoding satellite imagery, and \cite{Liu2018-pa} to forecasting wind speeds, showing not only the effectiveness but also the breadth of applications.
Although spectrograms maintain a lot of information, there still requires fine-grained hyperparameter tuning on the number of frequency bins, duration, and overlap. Moreover, dominate frequencies may not be known a priori, which can yield inaccurate or uninterpretable results.

Recent approaches to incorporate more meaningful filters have shown promising results in neural networks. \cite{Ravanelli2018-xi} propose a convolutional neural network based on parameterized $sinc$ functions to preform band-pass filtering in the first convolutional layer, where only the low and high cutoff frequencies are learned from the data. \cite{Rodriguez2020-zw} propose an adaptive wavelet neural network that extracts features using a multiresolution analysis approach for image classification. Their network learns wavelet coefficients by using a 2D lifting scheme and effectively reduces the number of trainable parameters. In this work we propose WaveNet, which builds upon these approaches, to learn parameterized wavelet functions using raw 1D signals. 





\section{Methodology}
\label{sec:methodology}

\subsection{Wavelet Transform}

A wavelet transform uses local wavelike functions to transform and present the signal's time-frequency information. Wavelets can be manipulated by moving to different locations on the signal or stretched and squeezed to cover different frequencies. The transform loosely quantifies the local matching of the wavelet and the underlying signal, whereby a large transform value is observed if the location and scale of the wavelet match the signal. Common wavelets used in practice include: Gaussian, Mexican hat (second derivative of a Gaussian), Haar, and Morlet. The integral wavelet transform is performed with an inner product of a function, $x(t)$, and mother wavelet, $\psi(t)$, as:
\begin{equation}
          W_x(s,\tau) = \int{x(t)\,\psi_{f,w}^*\Big(\frac{t - \tau}{s}\Big)\, dt}.
\end{equation}
For this work we use the complex Morlet wavelet with frequency, $f$, and width, $w$:
\begin{equation}
    \psi_{f,w}(t) = \underbrace{\big(s_t (2\pi)^{-\frac{1}{2}}\big)^{-\frac{1}{2}}}_{\substack{\text{Normalization}}} \underbrace{\exp{\big(\text{i}2\pi f t}\big)}_{\substack{\text{Complex} \\ \text{sinusoid}}} \underbrace{\exp{\Big(-\frac{t^2}{2s_t^2}}\Big)}_{\substack{\text{Gaussian} \\ \text{envelope}}},
\end{equation}
where $s_f = fw^{-1}$ and $s_t = (2\pi s_f)^{-1}$ are the standard deviation (resolution) of the wavelet transform in the spectral and in the temporal domains, respectively. The choice of the wave function is determined by the structural similarities with the data.

\subsection{Wavelet Network}
\label{sec:waveletnet}

WaveNet uses a wavelet transform as the first layer of a neural network where the convolution is a parameterized function of a mother wavelet. Specifically, we use $\psi_{f,w}(t)$ defined previously where $f$ and $w$ are trainable parameters learned through backpropagation. In the forward pass, the real and imaginary convolutional output are combined and the magnitude propagates to subsequent layers. Input signals are not standardized, thus batch normalization is applied following the transform layer to regularize the data and stabilize training. Thereafter, the values are input to zero or more fully-connected layers with $tanh$ nonlinearities before the final linear output layer.

Using too small a learning rate, $\eta$, during training will yield very small updates to the wavelet parameters. As such, different learning rates are set for $f$ and $w$ in the transform layer and all other parameters, $\theta$, in subsequent layers. Specifically,  
\begin{equation*}
    \begin{aligned}
        (f,w)^0_t &= (f,w)^0_{t-1} - \eta_0 \nabla_{(f,w)^0}LL(\theta)\\
        \theta^{1:}_t &= \theta^{1:}_{t-1} - \eta_1 \nabla_{\theta^{1:}}LL(\theta),
    \end{aligned}
\end{equation*}

where $\eta_0 = 0.1$ and $\eta_1 = 0.0001$. This change provides emphasis on the network to optimize for the wavelet parameters and significantly stabilizes training and improves performance. However, too large or an incorrect weight update to $f$ and $w$ can lead to negative or exceedingly large values, which is undesirable for the wavelet transform.
Therefore, after every mini-batch update, the values $f$ and $w$ are clipped between $[0.5, 30]$ and $[4, 15]$, respectively.


\section{Simplified Example}
\label{sec:simplified}
To demonstrate the efficacy of WaveNet we train a network to classify synthetic data composed of a background signal and time independent events of a predefined frequency. The intended result is to have the transform layer fit the event frequencies such that a single linear output layer can classify the samples. Data are generated for two classes using a sampling rate of $256$ Hz:
\begin{equation}
        y_{A,B}(t) = \sin(2\pi f_b t+\varphi_0) + \frac{1}{\sigma \sqrt{2\pi}}\exp{\Big(\frac{-(t-\mu)^2}{2\sigma^2}\Big)}\sin(2\pi f_{0,1}t+\varphi_1) + \varepsilon.
\end{equation}

The background signal is a sine wave with $f_b=9$ Hz and added noise, $\varepsilon$, from a normal distribution. Each class, A with $f_0=5$ Hz and B with $f_1=15$ Hz, is localized in time via a Gaussian envelope that is randomly shifted and centered by $\mu$ with $\sigma=0.8$. All data samples have a duration of \SI{0.80}{\second} (i.e. $205$ time steps) and are randomly phase shifted by $\varphi_0$ and $\varphi_1$ between $[0, 2\pi]$. Once generated, data are partitioned into training and test splits with equal class distributions, totaling $240$ and $60$ data samples, respectively.

A simple WaveNet is initialized with two wavelet filters having frequency values of $8$ and $12$ Hz and a linear output layer to discriminate between the two classes. A constant width value of $w=10$ is set as non trainable and used for both filters. The network optimizes the cross entropy loss on the training data using adaptive moment estimation. Within the first $60$ epochs, the network starts to achieve high classification accuracy; however, we continue training for a total of $700$ epochs to visualize how the wavelet parameters and network performance changes over time.

\begin{figure}[h]
    \begin{center}
        \includegraphics[width=0.7\textwidth]{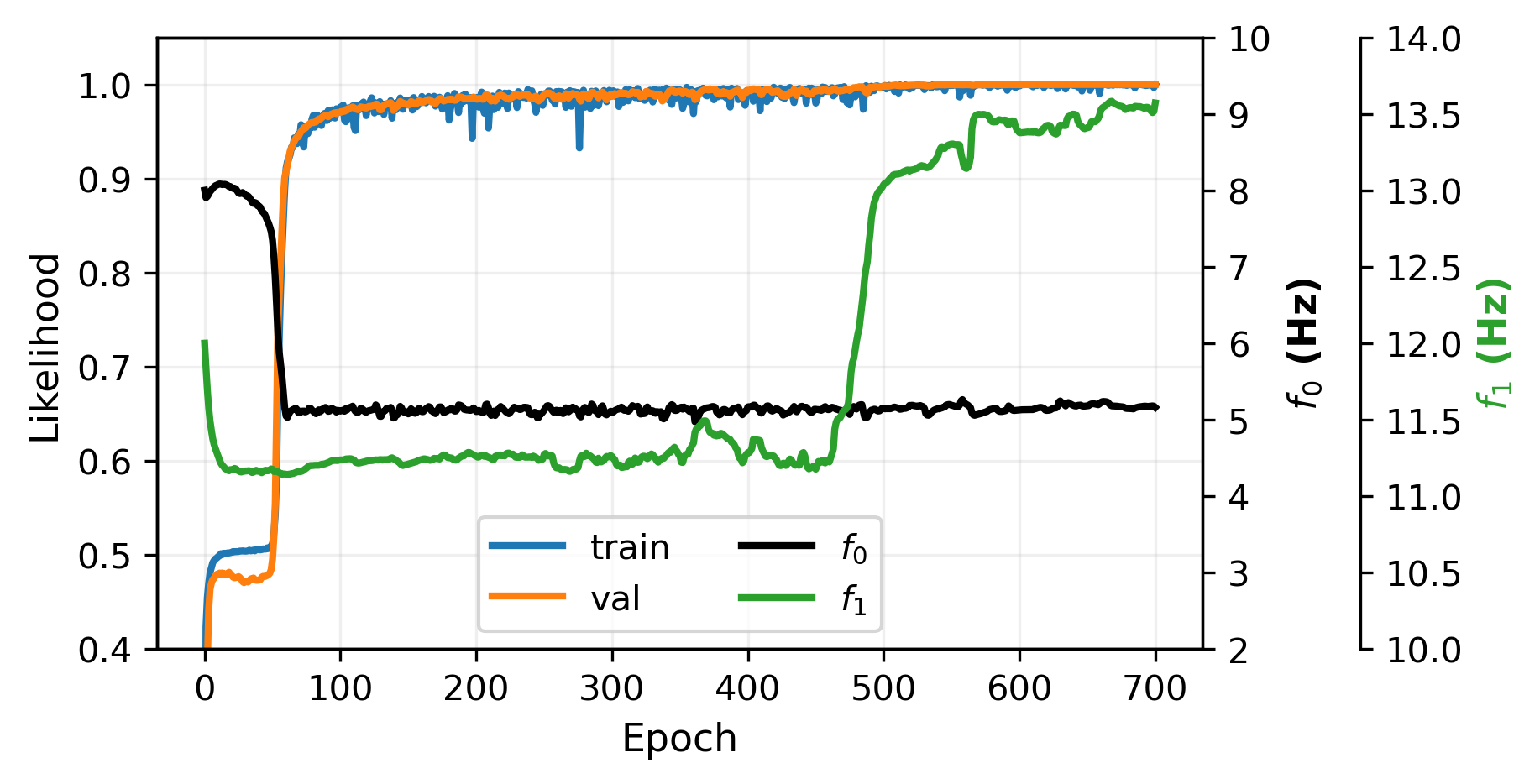}
    \end{center}
    \caption{Learning curves and central frequency values during training with simplified data.}
    \label{fig:simplified}
\end{figure}

\fref{fig:simplified} illustrates the likelihood during training as well as the change to $f_0$ and $f_1$. At $60$ epochs $f_0$ rapidly approach a value of $5$ HZ, matching the target event frequency for class A. As a result, the likelihood simultaneously rises to a value near one for both training and validation, yielding $98\%$ classification accuracy. With only two classes, the network performs well when matching a single event frequency, but the training loss continues to oscillate. After $480$ epochs we observe the training loss smooth out and the other wavelet frequency, $f_1$, approach the desired value of $15$ Hz and the network achieve $100\%$ test accuracy.

We conduct a second experiment with a standard fully-connected network of two hidden layers of $10$ units and nonlinear activations. Here, we use the same training and test partitions, and find the network to quickly overfit on the training data. Specifically, when noise, $\varepsilon$, is added to $y_{A,B}(t)$, the network severely lacks generalization and only achieves $45\%$ accuracy on the test data. Thus, making WaveNet more desirable, in terms of generalizability and learning performance, for data with quasiperiodic variations in time.

\section{Gravity Wave Identification}
\label{sec:gw}
Automatic detection of atmospheric gravity waves from satellite imagery is of interest to assimilate and improve the accuracy of numerical weather prediction models (\cite{Miller2015-hg}). Gravity waves are initiated by disturbances to the density structure of the atmosphere and restored via gravity and buoyant forces. In this work we show the use of WaveNet to classify a patch of satellite imagery as being either a gravity wave or cloud.

\subsection{Dataset Details}

Gravity wave and cloud data are acquired from the Cooperative Institute for Research in the Atmosphere (CIRA)'s Geostationary Operational Environmental Satellite (GOES)-16/17 Loop of the Day\footnote{\url{https://rammb.cira.colostate.edu/ramsdis/online/loop_of_the_day/}}. Each loop includes monochromatic imagery from animated GIFs (seven gravity wave and seven cloudy animations) with standard pixel-value intensities between $0...255$ (i.e., not true reflectance/brightness temperatures) standardized to have a max value of one. However, all animations are captured using the visible \SI{0.64}{\um} wavelength band from the Advanced Baseline Imager. The GOES-R series satellites have a nominal resolution of \SI{2.0}{\km}, but the animations are zoomed and cropped at different aspects with no specific map projection. 

\begin{figure}[h]
    \begin{center}
        \includegraphics[width=1.0\textwidth]{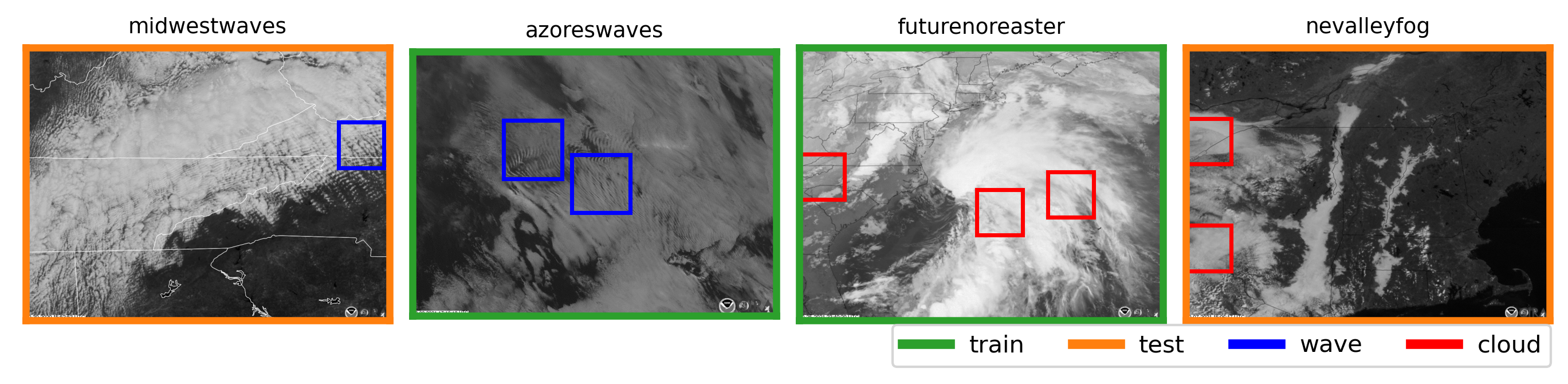}
    \end{center}
    \caption{A subset of satellite imagery frames with labeled gravity wave and cloudy scenes. Individual boxes represent the patch of the image used for the training and test datasets.}
    \label{fig:datasamples}
\end{figure}

For all $14$ animations, we hand select and label patches of size $128\times128$. Not every image within the animations actually contain gravity waves or clouds, and so identifying known patches a priori helps to create a more accurate dataset, although it is not representative of all wave or cloudy scenes. \fref{fig:datasamples} shows frames from a subset of animations used for training and testing with corresponding labels. Note, separate animations are held out for testing. We also shuffle and truncate each dataset to have an equal number of per class samples. 

Individual patches are preprocessed in such a way that intensity values over the spatial domain are candidate 1D samples. This differs from the traditional time-series data used in time-frequency analysis with wavelet transforms. However, we can treat the vectors of intensity values the same as time-series data with some predefined sampling rate. Vertical and horizontal vectors of pixel-intensities are cut from each patch for all $128$-$x$ and -$y$ coordinates and assigned the patch label. After slicing every patch, we are left with training data of size $165376\times1\times128$ and testing data of size $33792\times1\times128$.

\subsection{Training Details}

Referencing the details in \sref{sec:waveletnet}, we initialize WaveNet with $20$ wavelet filters composed of linearly spaced frequencies between $[1.5, 25]$ Hz and a single hidden layer of five nonlinear fully-connected units. The width parameter, $w$, is held constant at a value of $10$ and not learned during training. Through a grid search over possible hyperparameters, we find the network need only be trained for $10$ epochs with a batch size of $128$. 

We also train and evaluate other fairly standard neural network architectures, namely: a fully-connected network (fc-net), 1D convolutional network (conv-net), and filter-bank network (bank-net). As with WaveNet, the training parameters for each network are found through a preliminary hyperparameter search. We find that using two hidden layers with $20$ units each performs best for the fc-net. The conv-net is found to have two convolutional layers with the first layer having four filters and the second with eight, both with a kernel size of three, and separated by max pooling. Lastly, the bank-net is most similar to WaveNet in that we use a fully-connected network with one hidden layer of five units, but with preprocessed wavelet transformed samples from a filter-bank of $20$ wavelets that have linearly spaced frequencies between $[1.5, 25]$ Hz.

Each network is trained and evaluated on the test data $10$ times with different random weight initializations. This is to capture a more robust view of network performance and to alleviate the potential of settling in a local minima. Accuracy is the primary metric of interest due to having a binary classification task with an equal number samples belonging to each class. However, we do also report on precision and recall from the test data to bring insight to the predictions of both classes.

\section{Experimental Results}
\label{sec:gw-results}


\begin{figure}
\centering
\begin{minipage}{.36\textwidth}
  \centering
  \captionsetup{width=.9\linewidth}
  \includegraphics[width=1\linewidth]{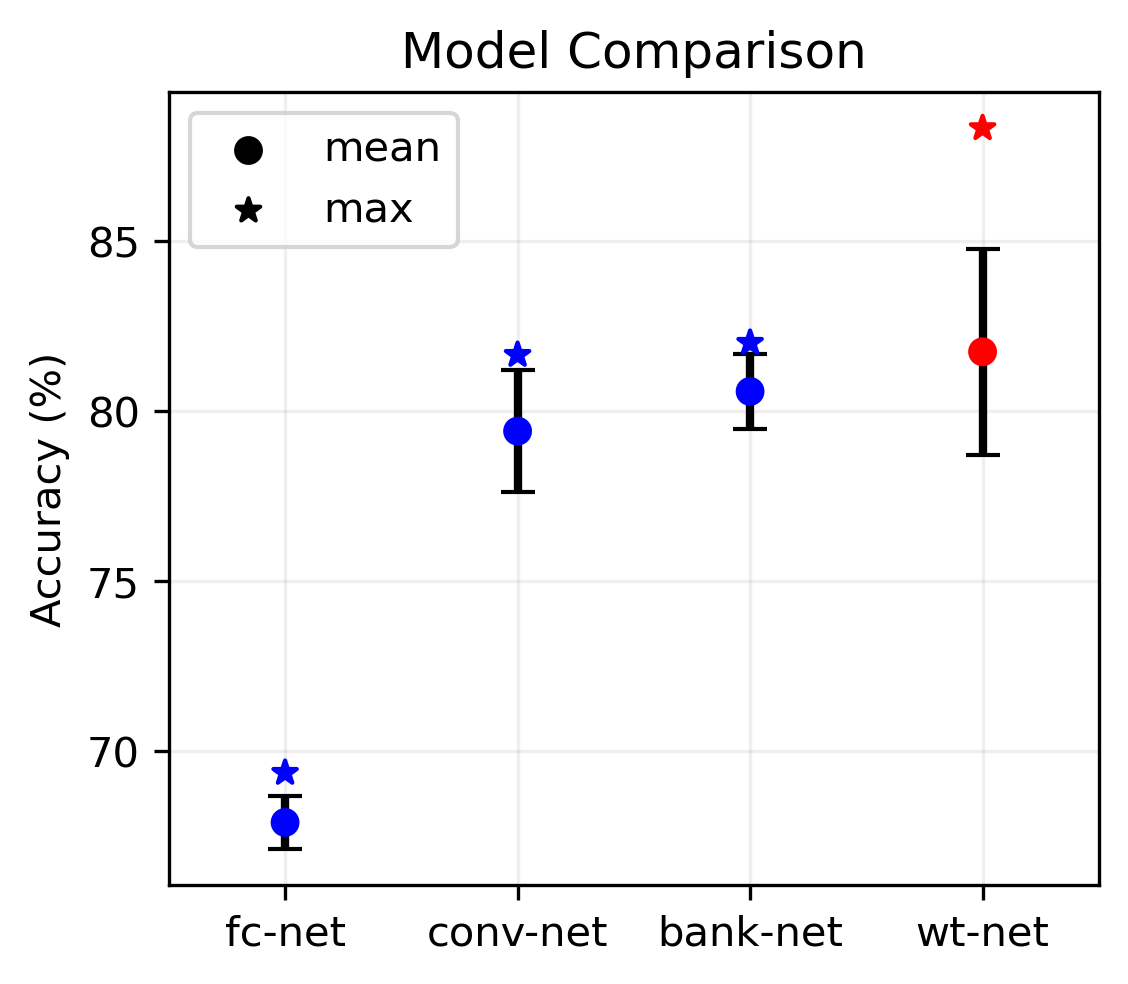}
  \caption{Test statistics of models trained on gravity wave data.}
  \label{fig:modelcomparison}
\end{minipage}%
\begin{minipage}{.64\textwidth}
  \centering
  \captionsetup{width=.9\linewidth}
  \includegraphics[width=1\linewidth]{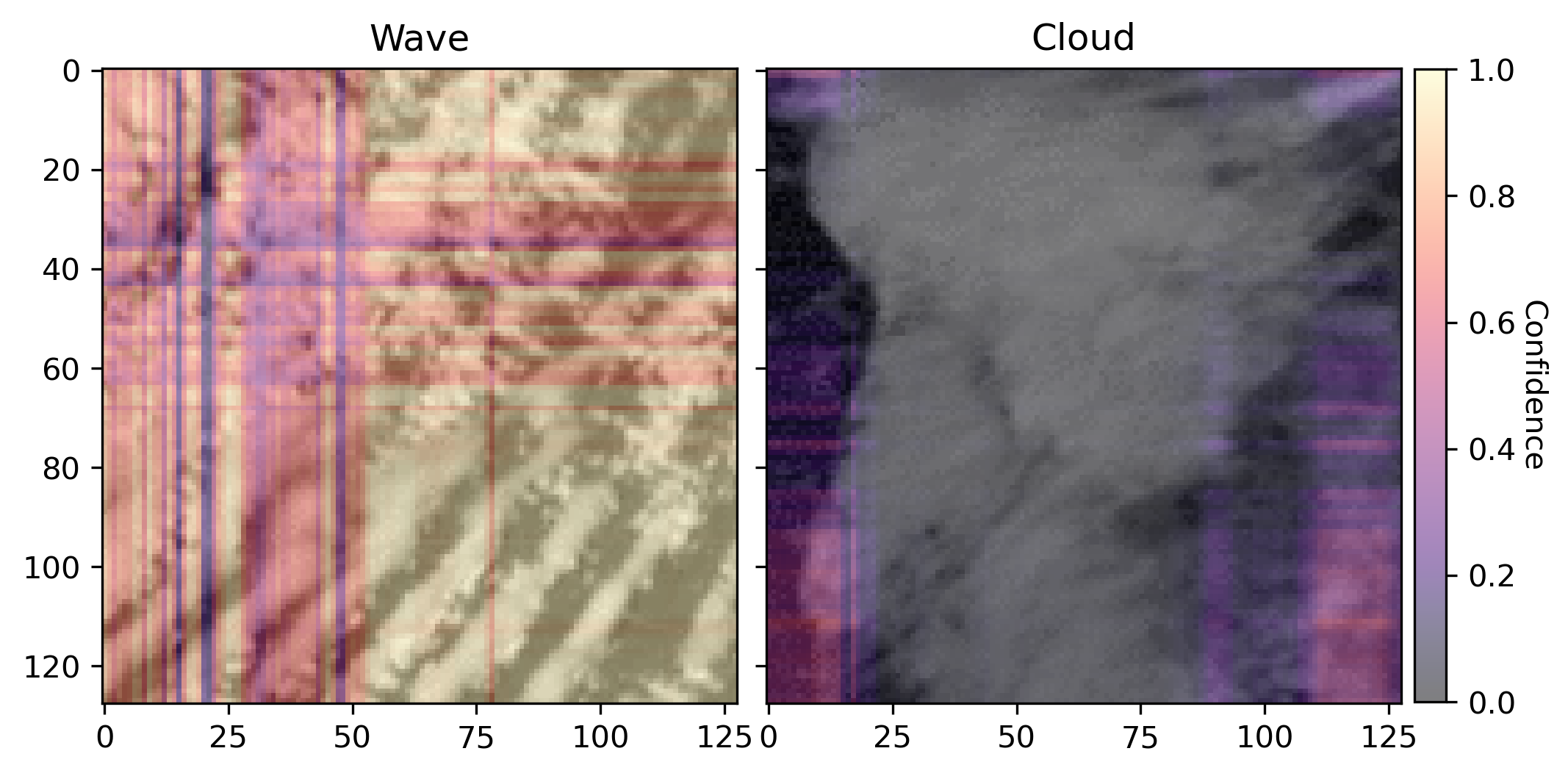}
  \caption{Confidence map on test wave and cloud patches. Higher values indicate greater confidence of a wave.}
  \label{fig:confidence}
\end{minipage}
\end{figure}

\fref{fig:modelcomparison} presents the aggregated test accuracies across all training trials for each network with the gravity wave dataset. The fc-net exhibits similar behavior to the simplified example discussed in \sref{sec:simplified}. Specifically, the network is prone to overfit on the training data after the first five epochs and performs the worst overall. This is likely due to the intrinsic noise and pixel-level variability in the data. If considering spatial context, and train the conv-net on the data, then we observe better generalization and an increase in accuracy of $12\%$ over fc-net. Adding additional fully-connected layers or increasing the depth of convolutions with the intent of identifying fine-grained patterns in the conv-net results in overfitting. Thus, it is most desirable to keep the networks relatively simple while capturing the spatial properties of the data. 

WaveNet and bank-net also exploit the spatial patterns, but rather through \textit{time}-frequency analysis with the wavelet transform, where time is actually in the spatial domain. The bank-net reaches a slightly higher accuracy, with a value of $80.57\pm1.12\%$, over the conv-net, but the improvement is not significant. The proposed WaveNet achieves the highest accuracy of $81.74\pm3.03\%$ and a maximum test accuracy of $88.32\%$. It is important to note that the WaveNet exhibits a large variability in performance with changes to the initial weights. This is as a result of training instability with the network occasionally showing relatively large oscillating loss values and the lack of generalization to unseen data (not shown within). However, the top performing network is of interest and studied in more detail from hereon.

Of the $16,896$ per class test samples, those labeled as a cloud have a higher recall value of $0.96$, whereas the gravity wave samples have a recall of $0.80$, indicating the occurrence of more false negatives. While this is seemingly undesirable, it can be reasoned as not every slice within a patch intersects with a gravity wave -- especially as animations evolve over time. We validate this claim by viewing the network's predictions over the entire patch. Recall that the network predicts on individual vertical and horizontal slices of an image patch. Therefore, to have a more comprehensive view of the prediction, we run a forward pass of every sample in a patch and compute a confidence map. Using the softmax (logistic) output for $P(y^{(i)}=1_{wave}\;|\;x^{(i)}; \Theta)$ at each sample, we calculate the outer product with the vector of all row and column values to produce a single matrix outlining the network confidence at each location. A pixel-value toward one indicates a greater probability of being labeled as a gravity wave. 

\fref{fig:confidence} overlays a confidence map on a single test patch of gravity wave and cloud images. In the lower-right quadrant of the gravity wave map, we see high confidence in the network's prediction of gravity waves. The least confident region is in the upper-left quadrant, which has high cloud coverage and no visible gravity waves. The cloud image has a max confidence of $0.62$ around the boarders where there are slices with both cloud and land mass visible. However, there are no wave-like slices that yield a high confidence value.

As for the transform layer within WaveNet, there are multiple filters that converge on nearly the same frequency values, i.e. $2.4$, $13.2$, and $29.5$ Hz. The majority of low-frequency values (between $1.5$-$7.0$ Hz) do not move far from their initial value, which we speculate to be representative in the original dataset. Reducing the number of filters could enable the network to learn more unique frequency values, but we find the additional filters to improve performance. 

\section{Conclusion}
\label{sec:conclusion}
The proposed WaveNet is a step toward interpretable-by-design networks that learns dominate frequencies in the data. Very few trainable parameters are required as with a parameterized function we effectively reduce the number of parameters to learn in the convolutional filter to two (i.e. frequency, $f$, and width, $w$). We demonstrate the efficacy of the network with a simplified dataset, showing how the network fits to the underlying frequencies. Additionally, we explore gravity wave classification and achieve $88.32\%$ with a handcrafted dataset of satellite imagery. This work can be further extended with different wavelet functions and other network architectures that may better suit the underlying data. Lastly, we believe there can be more transparency by evaluating post-hoc explainability methods to bring a greater understanding to what wavelet parameters are most useful for class or value predictions.
\subsubsection*{Acknowledgments}
This work is part of the NSF AI Institute for Research on Trustworthy AI in Weather, Climate, and Coastal Oceanography (AI2ES) and is supported by the NSF under Grant No. 2019758.

\bibliography{iclr2022_conference}
\bibliographystyle{iclr2022_conference}


\end{document}